\documentclass[runningheads]{llncs}

\usepackage[T1]{fontenc}
\usepackage{graphicx}
\usepackage{booktabs}
\usepackage{multirow}
\usepackage{amsmath}
\usepackage{amssymb}
\usepackage{url}
\usepackage{float}
\usepackage{placeins}
\usepackage{marvosym}

\begin{document}

\title{Learning from Acquisition: Metadata-driven Multimodal Pre-training for Cardiac MRI}
\titlerunning{MetaCLIP-CMR}

\author{ Xueyi Fu\inst{1} \and Liwei Hu\inst{2} \and Zi Wang\inst{2}\textsuperscript{\Letter} \and Guang Yang\inst{2,3,4,5}\textsuperscript{\Letter} }

\authorrunning{X. Fu et al.}

\institute{ Department of Surgery \& Cancer, Imperial College London, London, UK \and Bioengineering Department and Imperial-X, Imperial College London, London, UK \and National Heart and Lung Institute, Imperial College London, London, UK \and Cardiovascular Research Centre, Royal Brompton Hospital, London, UK \and School of Biomedical Engineering \& Imaging Sciences, King's College London, London, UK\\\textsuperscript{\Letter}
\email{\{zi.wang,g.yang\}@imperial.ac.uk}
 }
\maketitle

\begin{abstract}
Cardiac magnetic resonance imaging (CMR) routinely records structured acquisition metadata, yet most CMR foundation models rely primarily on image-only pre-training and leave this naturally available source of weak semantic supervision largely underexplored. We propose MetaCLIP-CMR, a metadata-driven framework based on Contrastive Language--Image Pre-training (CLIP), which converts imaging modality, anatomical view, scanner vendor, field strength, and scanner model into textual supervision for CMR representation learning. The pretrained image encoder is evaluated on imaging modality classification, cine view classification, and cardiac segmentation. MetaCLIP-CMR achieves 86.8\% modality accuracy and 86.5\% cine view accuracy, clearly outperforming ImageNet and masked reconstruction initialisations. For downstream cardiac segmentation, MetaCLIP-CMR consistently obtains the highest Dice score across the evaluated ACDC and M\&Ms cine short-axis (SAX) settings under both full-data and 20\% fine-tuning regimes. Compared with recent image-focused large-scale CMR pre-training models, MetaCLIP-CMR achieves comparable ACDC segmentation performance, while requiring less than 1\% of their pre-training image scale. These results suggest that metadata learning offers a natural and easy-to-use strategy for transforming routinely recorded acquisition information into effective supervision for foundation-level CMR representation learning, highlighting the promise of metadata-driven multimodal pre-training.

\end{abstract}

\section{Introduction}

Cardiac magnetic resonance imaging (CMR) provides complementary anatomical,
functional, and tissue-characterisation information across cine, late
gadolinium enhancement (LGE), mapping, flow, and other acquisition sequences
\cite{Multi-modality,Present,CMRxRecon2024}. This diversity also introduces substantial
variation in image appearance across modalities, anatomical views, scanner
vendors, and field strengths.

Recent medical and CMR foundation models have mainly relied on image-only
self-supervised learning, including masked image modelling, contrastive
learning, and self-distillation
\cite{Masked,Foundation,Afoundation}. In CMR, the DINO-based model
\cite{Towardsacardiovascular} learns visual representations from large-scale
heterogeneous CMR data, while CineMA~\cite{Fu2026} applies masked image
modelling to large-scale cine CMR. These studies show that CMR-specific
pre-training can support downstream segmentation and classification
\cite{Towardsacardiovascular,Towardscardiac}. However, their supervision is
derived primarily from image appearance and does not explicitly incorporate
acquisition-level information such as imaging modality, anatomical view, vendor, or
field strength.

Vision--language models such as CLIP~\cite{LTVM} introduce semantic
supervision by aligning images with text. Medical variants have further
demonstrated the value of image--text pre-training for transferable medical
representations \cite{BiomedCLIP,CLIP}. Their application to CMR is limited by
the availability of paired clinical reports or manually curated textual
annotations. In contrast, acquisition metadata are routinely recorded during
scanning and provide structured information without additional manual
annotation.

We therefore propose MetaCLIP-CMR, a metadata-driven image--text pre-training
framework that converts imaging modality, anatomical view, scanner vendor, field
strength, and scanner model into textual descriptions. The pretrained image
encoder is evaluated on modality classification, cine view classification,
and cine short-axis segmentation. The classification tasks assess whether
metadata-defined concepts are transferred to the image representation, while
segmentation evaluates transfer to a pixel-level task not directly specified
by the text. Beyond internal comparisons with ImageNet and masked reconstruction
initialisation, we further contextualise MetaCLIP-CMR against recent
image-focused CMR pre-training models to assess the efficiency of
metadata-driven pre-training relative to large-scale image-only representation
learning.

\textbf{Our main contributions are:}
\begin{itemize}
\item We introduce a metadata-driven CMR image--text pre-training framework
that transforms routinely recorded acquisition information into natural,
structured, and easy-to-use weak semantic supervision, without requiring
manually paired clinical reports or curated text annotations.
\item We demonstrate that MetaCLIP-CMR learns modality- and view-aware
representations that transfer consistently to cine short-axis segmentation,
outperforming ImageNet and masked reconstruction baselines in imaging classification
and cardiac segmentation.
\item Compared with recent image-focused large-scale CMR pre-training
models, MetaCLIP-CMR reaches comparable segmentation performance
while using less than 1\% of their pre-training image scale, highlighting
the potential of metadata learning for efficient CMR representation
pre-training.
\end{itemize}

\section{Method}

\subsection{Overview}
\begin{figure*}[h]
    \centering
    \includegraphics[width=\textwidth]{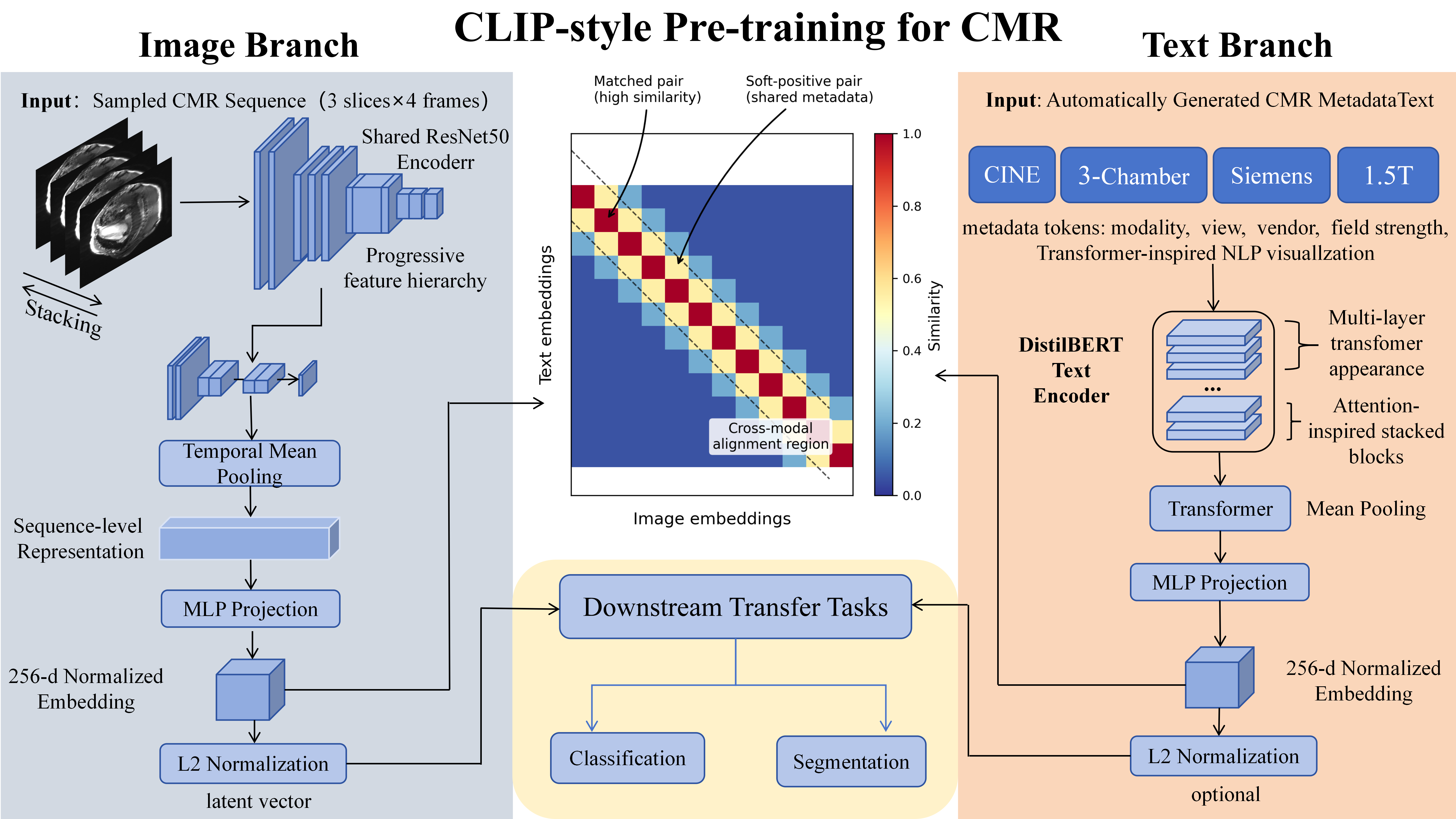}
    \caption{
    Overview of MetaCLIP-CMR. Image sequences and metadata-derived text are
    projected into a shared 256-dimensional space and aligned using the proposed
    soft-label contrastive loss.
    }
    \label{fig:framework}
\end{figure*}
MetaCLIP-CMR aligns CMR sequences with metadata-derived text in a shared
embedding space. As shown in Fig.~\ref{fig:framework}, the model contains image
and text encoders with projection heads trained using a metadata-aware
contrastive loss. After pre-training, only the image encoder is retained.

\subsection{Preprocessing and Metadata Text}

\begin{figure}[t]
    \centering
    \includegraphics[width=\linewidth]{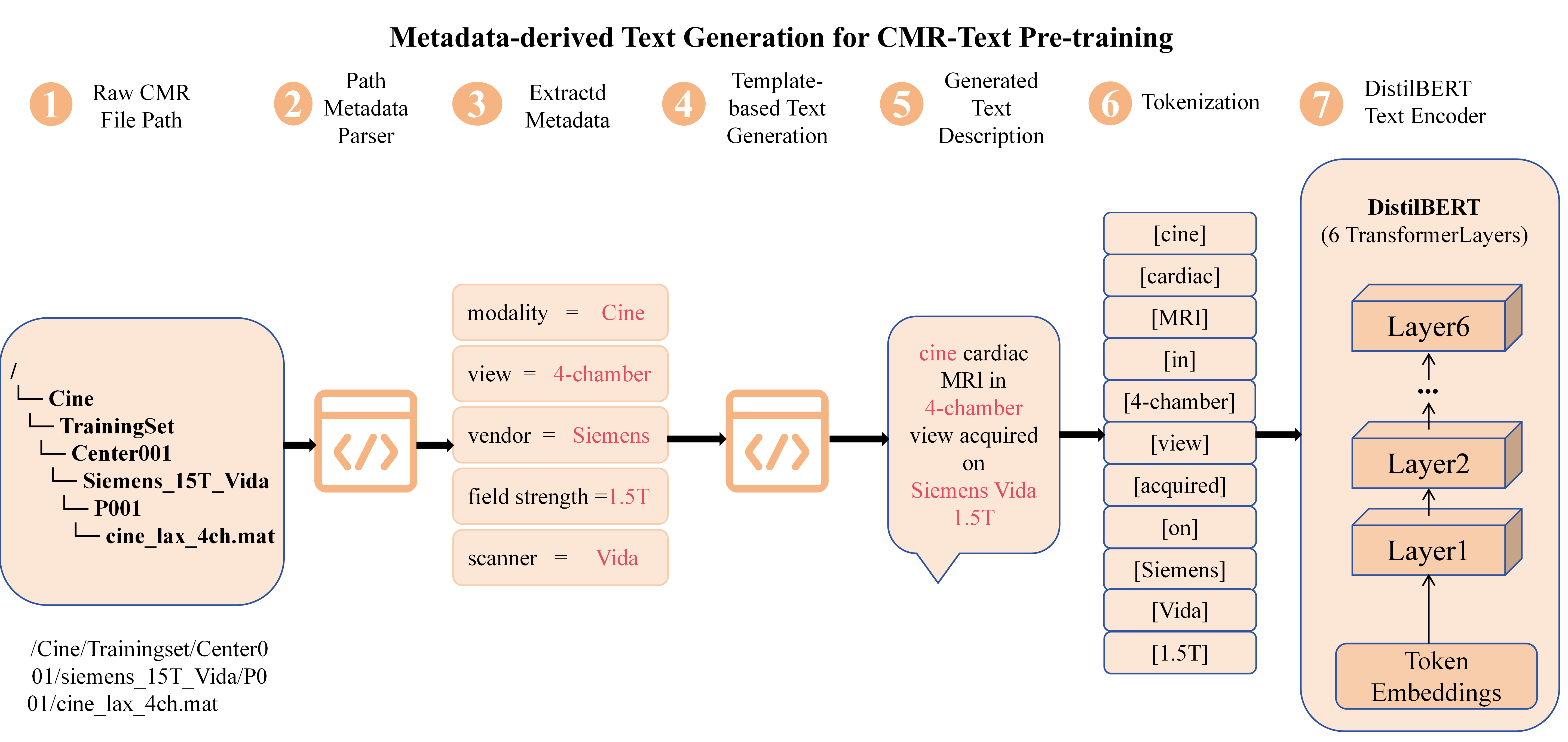}
    \caption{
    Metadata-derived text generation from CMR modality, anatomical view, scanner
    vendor, field strength, and scanner model.
    }
    \label{fig:metadata_generation}
\end{figure}

\noindent\textbf{Data preprocessing.}
Raw k-space data from MMCMR-427K database~\cite{Enablingultrafastcardiovascularimaging} are reconstructed offline using sum-of-squares
(SOS) reconstruction and stored as 2D \texttt{.npy} slices. Each slice is
z-score normalised and resized to $224\times224$ pixels. During pre-training,
we sample $N_s=3$ slices and up to $N_f=4$ frames per sequence, yielding up
to 12 images. If fewer than four frames are available, all available frames
are used without repetition. Augmentation includes random horizontal flipping,
rotation ($\pm10^\circ$), translation of up to 10\%, and Gaussian noise
($\sigma\in[0.005,0.05]$), with consistent spatial transformations across
images from the same sequence.

\noindent\textbf{Metadata text generation.}
Structured acquisition metadata are converted into text using predefined
templates (Fig.~\ref{fig:metadata_generation}). For sequence $i$,
\begin{equation}
t_i=\mathcal{T}(a_{i1},\ldots,a_{iK}),
\end{equation}
where the attributes include CMR modality, anatomical view, scanner vendor, field
strength, and scanner model. An example is:
\textit{``cine cardiac MRI in short-axis view acquired on a 3.0T scanner''}.
The descriptions are generated automatically from metadata parsed from the
dataset organisation, without manual annotation.

\subsection{Multimodal Pre-training}

The image encoder $f_{\theta}$ is a single-channel ResNet-50 that processes
each sampled image independently. Let $x_{it}$ denote the $t$-th sampled image
from sequence $i$. The corresponding image feature is
\begin{equation}
h_{it}=f_{\theta}(x_{it})\in\mathbb{R}^{2048}.
\end{equation}
The features from the $T_i$ sampled images are mean-pooled to obtain a
sequence-level representation:
\begin{equation}
\bar{h}_i=\frac{1}{T_i}\sum_{t=1}^{T_i}h_{it}.
\end{equation}
The text encoder $g_{\phi}$ is DistilBERT
\cite{sanh2020distilbertdistilledversionbert}. Image and text representations
are projected into a shared 256-dimensional space by $W_v$ and $W_u$,
respectively, and L2-normalised:
\begin{equation}
v_i=
\frac{W_v\bar{h}_i}
{\|W_v\bar{h}_i\|},
\qquad
u_i=
\frac{W_u g_{\phi}(t_i)}
{\|W_u g_{\phi}(t_i)\|}.
\end{equation}

Following CLIP~\cite{LTVM}, the image--text similarity matrix is
\begin{equation}
S_{ij}=v_i^\top u_j.
\end{equation}
Standard CLIP treats only the matched image--text pair as positive and all other pairs in the batch as negatives. In CMR, however, different sequences may share acquisition semantics such as imaging modality or anatomical view. We therefore assign weaker positive weights to non-matched samples sharing these attributes.

To account for acquisition-level similarities between CMR sequences, we replace the one-hot CLIP targets with metadata-aware soft targets:
\begin{equation}
\tilde{Q}_{ij}
=
\mathbb{I}(i=j)
+
\alpha\,\mathbb{I}(i\neq j,\,m_i=m_j)
+
\beta\,\mathbb{I}(i\neq j,\,r_i=r_j),
\end{equation}
where $m_i$ and $r_i$ denote the modality and anatomical view of sequence
$i$. The exact image--text pair receives weight 1, while non-matched samples
sharing the same modality or view receive weaker positive weights
$\alpha=\beta=0.05$. If both attributes are shared, the two contributions are
added. The view term is applied only when valid view annotations are
available.

The targets are normalised along each row:
\begin{equation}
Q_{ij}
=
\frac{\tilde{Q}_{ij}}
{\sum_{k=1}^{B}\tilde{Q}_{ik}}.
\end{equation}
Thus, each row defines a probability distribution over the candidate samples
in a batch of size $B$.

The bidirectional soft-label contrastive loss is
\begin{equation}
\mathcal{L}
=
\frac{1}{2}
\left[
\mathrm{SCE}(Q,S/\tau)
+
\mathrm{SCE}(Q,S^\top/\tau)
\right],
\end{equation}
where $\tau$ is a learnable temperature and
\begin{equation}
\mathrm{SCE}(Q,Z)
=
-\frac{1}{B}
\sum_{i=1}^{B}\sum_{j=1}^{B}
Q_{ij}
\log
\frac{\exp(Z_{ij})}
{\sum_{k=1}^{B}\exp(Z_{ik})}.
\end{equation}
The row-wise normalisation corresponds to selecting texts for each image in
the image-to-text direction and images for each text in the text-to-image
direction.

For image-only comparison, we train a ResNet-50 masked reconstruction baseline
following the general masked image modelling paradigm~\cite{Masked}. A
lightweight decoder reconstructs masked regions and is discarded after
pre-training.

\subsection{Downstream Transfer}

The pretrained image encoder is evaluated on CMR modality classification, cine
view classification, and cardiac segmentation.

We use a two-stage linear probing and fine-tuning protocol (Probe+FT). During the
probe stage, the image encoder is frozen and only the task-specific head or
decoder is trained. The encoder is then unfrozen and the full network is
fine-tuned using a smaller learning rate for the encoder.

\section{Experiments and Results}

\subsection{Datasets and Implementation}

\begin{table}[h]
\centering
\caption{
Datasets used for pre-training and downstream evaluation. Our used MMCMR-427K subset
contains 10 CMR modalities from five vendors and 1.5T/3.0T systems.
}
\label{tab:dataset}
\small
\begin{tabular}{lll}
\toprule
Dataset & Task & Train / Test \\
\midrule
MMCMR-427K & Pre-training & 4,672 / 2,385 sequences \\
           & Modality classification & 4,672 / 1,914 sequences \\
           & Cine view classification & 4,672 / 576 sequences \\
\midrule
ACDC        & Segmentation
& 100 / 50 cases \\
M\&Ms       & Segmentation
& 150 / 136 cases \\
\bottomrule
\end{tabular}
\end{table}
In this study, We use a subset of the original MMCMR-427K database~\cite{Enablingultrafastcardiovascularimaging}. The source database contains 427,465 paired k-space samples. The subset used in this work contains 7,057 CMR sequences spanning 10 CMR modalities, five vendors, and 1.5T/3.0T systems, with 4,672 used for pre-training and 2,385 held out (Table~\ref{tab:dataset}). Held-out MMCMR-427K sequences are used for CMR modality and cine view
classification. ACDC~\cite{bernard2018acdc} and
M\&Ms~\cite{campello2021multi} are used for cine short-axis segmentation. The evaluation sets remain unchanged, and the
same reduced-data subsets are used for all initialisation methods.

MetaCLIP-CMR uses a ResNet-50 image encoder
\cite{he2015deepresiduallearningimage} and a DistilBERT text encoder
\cite{sanh2020distilbertdistilledversionbert}, with both branches projected
into a 256-dimensional embedding space. Pre-training is performed for
30 epochs using AdamW
\cite{loshchilov2019decoupledweightdecayregularization}. We compare
MetaCLIP-CMR with ImageNet initialisation and image-only masked
reconstruction. Classification is evaluated by accuracy and segmentation by
case-level 3D Dice.

\subsection{Metadata-aligned Classification}

To assess whether metadata-derived supervision is transferred to the image
representation, we evaluate CMR modality and cine view classification, as both
attributes are explicitly encoded in the pre-training text.

MetaCLIP-CMR achieves the highest accuracy in both tasks
(Table~\ref{tab:cls}). Its strong probe performance, together with the small additional gain after fine-tuning, indicates that modality- and view-level semantics are already linearly accessible from the frozen representation. 

This suggests that metadata-driven pre-training does not merely improve downstream
classification accuracy, but organises the image embedding space around
clinically meaningful acquisition concepts. These results provide direct
evidence that routinely recorded acquisition metadata can act as effective
semantic supervision for learning acquisition-aware CMR representations.


\begin{table}[h]
\centering
\caption{
CMR modality and cine view classification accuracy. Probe denotes frozen-encoder
evaluation, and Probe+FT denotes subsequent end-to-end fine-tuning.
}
\label{tab:cls}
\small
\begin{tabular}{lcccc}
\toprule
Method
& Mod. Probe
& Mod. Probe+FT
& View Probe
& View Probe+FT \\
\midrule
ImageNet
& 0.686 & 0.751 & 0.740 & 0.793 \\
Masked recon.
& 0.658 & 0.729 & 0.655 & 0.682 \\
MetaCLIP-CMR
& \textbf{0.864}
& \textbf{0.868}
& \textbf{0.830}
& \textbf{0.865} \\
\bottomrule
\end{tabular}
\end{table}

\subsection{Transfer to Cine SAX Segmentation}

To assess transfer beyond metadata-aligned classification, we evaluate cine
short-axis segmentation on ACDC and M\&Ms using the same Probe+FT protocol
for all pre-training methods.

\begin{figure*}[t]
    \centering
    \includegraphics[width=\textwidth]{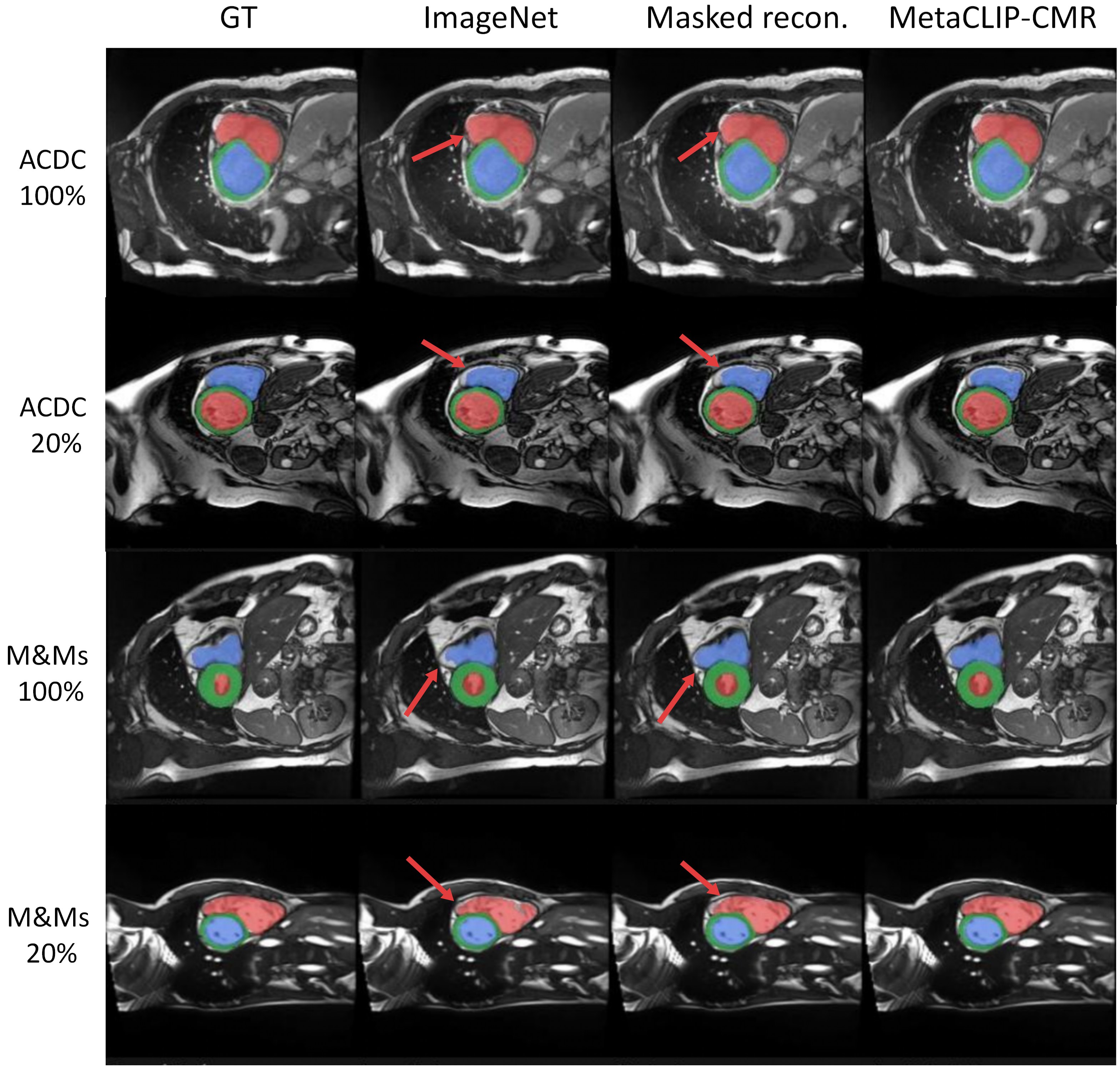}
    \caption{
    Representative cine SAX segmentation results on ACDC and M\&Ms under
    full-data and 20\% settings. RV, myocardium (MYO), and LV are shown in
    red, green, and blue, respectively.
    }
    \label{fig:seg_examples}
\end{figure*}

MetaCLIP-CMR obtains the highest Dice across all four evaluated settings
(Table~\ref{tab:seg}) and produces the most accurate visual segmentation among
the compared pre-training methods (Figure~\ref{fig:seg_examples}). The improvement is
particularly evident in the 20\% ACDC and full-data M\&Ms settings, suggesting
that metadata-driven pre-training can provide useful transferable representations
even for pixel-level tasks that are not directly specified by the pre-training
text. Figure~\ref{fig:seg_examples} shows representative ACDC and M\&Ms results.
The arrows indicate local contour errors in the ImageNet and masked
reconstruction predictions.

Together with the classification results, these findings indicate that
acquisition metadata provide a natural and effective supervision signal for
learning CMR representations that transfer across both acquisition-aware and
anatomy-oriented downstream tasks.

\begin{table}[h]
\centering
\caption{
Case-level mean Dice for cine SAX segmentation. The 100\% settings use
100 ACDC and 150 M\&Ms training cases, while the 20\% settings use
20 and 30 cases, respectively.
}
\label{tab:seg}
\small
\begin{tabular}{lcccc}
\toprule
Method
& ACDC 100\%
& ACDC 20\%
& M\&Ms 100\%
& M\&Ms 20\% \\
\midrule
ImageNet
& 0.8997 & 0.8642 & 0.8260 & 0.7965 \\
Masked recon.
& 0.8994 & 0.8635 & 0.8300 & 0.7966 \\
MetaCLIP-CMR
& \textbf{0.9021}
& \textbf{0.8795}
& \textbf{0.8368}
& \textbf{0.8008} \\
\bottomrule
\end{tabular}
\end{table}
\subsection{Comparison with Large-scale CMR Pre-training Models}

Here, we further contextualise MetaCLIP-CMR against recent image-focused CMR
large-scale pre-training models on ACDC segmentation. Although differences in data,
architectures, and evaluation protocols prevent a strictly controlled
comparison, the reported results provide a useful reference for downstream
performance.

MetaCLIP-CMR achieves a mean ACDC Dice of 0.902, close to the DINO-based CMR
model (0.906) \cite{Towardsacardiovascular} and CineMA (0.909) \cite{Fu2026}. The latter models were pretrained on approximately 36 million CMR images
\cite{Towardsacardiovascular} and more than 15 million cine CMR images
\cite{Fu2026}, respectively, whereas MetaCLIP-CMR uses approximately 0.2 million but more diverse CMR images~\cite{Enablingultrafastcardiovascularimaging}. This suggests that metadata-driven multimodal pre-training can approach the performance of much larger image-focused CMR models with less than 1\% of their pre-training image scale.


\section{Conclusion}

We presented MetaCLIP-CMR, a metadata-driven image--text pre-training
framework. By converting routinely recorded acquisition
information into textual descriptions, MetaCLIP-CMR turns a
naturally available and easy-to-use source of weak semantic information into
supervision for CMR representation learning.


Future work will extend metadata-driven pre-training with richer protocol
information and temporal modelling across larger multi-centre, multi-vendor, and multi-disease CMR cohorts, while exploring broader downstream tasks such as cardiac phenotyping, disease classification, image quality control, and cross-modality retrieval.

\clearpage
\bibliographystyle{splncs04}
\bibliography{references}

\end{document}